\def\year{2018}\relax
\documentclass[letterpaper]{article} 

\usepackage{amsfonts,amsmath,algorithmic}
\usepackage{tablefootnote}
\usepackage{caption}
\usepackage{subcaption}

\usepackage{aaai18}  
\usepackage{times}  
\usepackage{helvet}  
\usepackage{courier}  
\usepackage{url}  
\usepackage{graphicx}  
\long\def\ignore#1{}
\newcommand{\vek}[1]{{\bf {#1}}}
\newcommand{\vx}{{\vek{x}}}

\newcommand{\vw}{{\vek{w}}}
\newcommand{\vh}{{\vek{h}}}
\newcommand{\vM}{{\vek{v}}}

\newcommand{\util}{{\alpha}}
\newcommand{\lmm}{{s}}

\newcommand{\vbeta}{{\boldsymbol{\beta}}}
\newcommand{\argmax}{{\text{argmax}}}
\newcommand{\argmin}{{\text{argmin}}}
\newcommand{\loss}{{\text loss}}

\RequirePackage[textsize=scriptsize]{todonotes}
\definecolor{turquoise}{cmyk}{0.65,0,0.1,0.1}
\definecolor{purple}{rgb}{0.85,0,2.85}
\definecolor{dark_green}{rgb}{0, 0.5, 0}
\definecolor{orange}{rgb}{0.8, 0.6, 0.2}
\definecolor{brown}{rgb}{0.5, 0.16, 0.16}

\newcommand{\rulesep}{\unskip\ \vrule\ }

\begin{document}
%
\title{Labeled Memory Networks for Online Model Adaptation}
\author{Shiv Shankar\\
shiv\_shankar@iitb.ac.in\\
IIT Bombay
\And
Sunita Sarawagi\\
sunita@iitb.ac.in\\
IIT Bombay
}

\maketitle
\begin{abstract}
Augmenting a neural network with memory that can grow without growing the number of trained parameters is a recent powerful concept with many exciting applications.  In this paper, we establish their potential in online adapting a batch trained neural network to  domain-relevant labeled data at deployment time.

We present the design of Labeled Memory Network (LMN), a new  memory augmented neural network (MANN) for fast online model adaptation.  We highlight three key features of LMNs.
First, LMNs treat memory as a second boosted stage following the trained network thereby allowing the memory and network to play complementary roles.  Unlike all existing MANNs that write to memory at every cycle, LMNs provide better memory utilization by writing only labeled data with non-zero loss.
Second, LMNs organize the memory with the discrete class label as the primary key unlike existing MANNs where key is a real vector derived from the input. This simple, yet surprisingly unexplored alternative organization, safeguards against catastrophic forgetting of rare labels that current LRU based MANNs are subject to. Finally, LMNs  model the evolving expertise of memory and network using a RNN, to determine online their respective weights

We evaluate online model adaptation strategies on five sequence prediction tasks, an image classification task, and two language modeling tasks. We show that LMNs are better than other MANNs designed for meta-learning. We also found them to be more accurate and faster than state-of-the-art methods of retuning model parameters for adapting to domain-specific labeled data.
\end{abstract}



\section{Introduction}

While deep learning models achieve impressive accuracy in supervised learning tasks such as computer vision~\cite{Krizhevsky_imagenetclassification},  translation~\cite{WuGNMT16}, and speech recognition~\cite{Yu14ASR} training such models places significant demands on the  amount of labeled data, computational resources, and manual efforts in tuning.  Training is thus considered an infrequently performed resource-intensive batch process.  
Many applications require trained models to quickly adapt to new examples and settings during deployment. Consider the online sequence prediction task that arises in applications such as text auto-completion, user trajectory prediction, or next-url prediction.  Even when the prediction model has been trained on many sequences, next-token predictions on new sequences can benefit from true tokens observed online.
Another example is image recognition where a trained model should be able to recognize new objects not seen during batch training based on a few examples. This has led to a surge of interest in  few-shot learning ~\cite{kaiser2017,SantoroBBWL16,VinyalsBLKW16,zemel17}. Few-shot learning is an artificially simplified setting where a small set of new labels define independent classification tasks. We consider the more useful but more difficult and less explored task where an existing image classifier has to be extended to handle new labels.

%
%

An established technique for model adaptation is to retune part or all of the model parameters using the domain-labeled data in a separate adaptation phase~\cite{daume2007}.  More recently, deep meta-networks have been proposed that "learn to learn" such adaptation~\cite{Rei15,Finn2017ModelAgnosticMF,Huang2015MaximumAP,ravi17}. Parameter retuning methods typically require a one-time adaptation step, and operating them in online settings is slow.
Secondly with a pre-trained network the multiple gradient updates of meta-learning tend to destroy useful information learned by the PCN. This phenomenon is closely related to 'catastrophic forgetting'  \cite{French99,KirkpatrickPRVD16}. 

An alternative to parameter retuning is memorizing.  In this approach neural networks are augmented  with memory that can grow without correspondingly increasing the number of parameters to be trained.  Many exciting uses have been found of such MANNs including program learning~\cite{GravesNTM}, question answering~\cite{Weston16,GulcehreCCB16}, learning rare events~\cite{kaiser2017}, and meta learning~\cite{SantoroBBWL16}.  They hold promise for model adaptation because memory they can cut short the conventional path of iterative training to percolate new facts to model parameters.  


However, our initial attempts at using existing MANNs like NTMs~\cite{GravesNTM} and DNTMs~\cite{GulcehreCCB16} for online sequence prediction did not improve the baseline model. One major challenge is correctly balancing the roles of the memory and batch-trained model.   On the one hand, we have a shared model trained over several instances, and on the other hand we have the few but more relevant instances encountered during testing. Recent MANNs designed for meta-learning~\cite{SantoroBBWL16} partition their roles by using the shared model to learn an embedding and the memory to implement a nearest neighbor like classifier. While this architecture works for few-shot learning where all labels are new, they cannot adapt classification models with softmax layers over a large shared label space.

\paragraph{Contributions}
In this paper we present a new MANN called Labeled Memory Network (LMN) that provides an easy, fast, and plug-and-play solution for adapting pre-trained models.  We highlight three design decisions that made LMNs suitable for such adaptation.

%



First, we apply ideas from boosting~\cite{Schapire:1999adaboost} and treat memory as a second-stage classifier that is updated only when the current loss is non-zero. Existing MANNs write to memory during every pass.  Even when the goal of the model is to use the memory to remember rare events \cite{kaiser2017}, the memory stores all events not just the rare ones. This causes a lot of memory to be wasted in storing non-rare vectors often displacing the rare ones.

Second, we propose a 'labeled memory' where the primary means of addressing a memory cell is by a class label.  This is in contrast to all existing MANNs that use a controller to generate a hidden vector of the input as key.  This simple, yet surprisingly unexplored alternative organization\footnote{Our method of using labels for addressing is very different from the discrete addressing mechanism proposed in \cite{GulcehreCCB16}.  Even though both result in discrete addressing, in our case the key is the label whereas in \cite{GulcehreCCB16} the key continues to be computed from input $\vx$.} has two advantages: First, the controller is freed from the vaguely supervised task of generating keys that are distinctive and relevant leading to better memory use.  Second, it safeguards against catastrophic forgetting of rare labels that current LRU based MANNs are subject to.

Third, we use the power of a RNN to adaptively define the roles of the memory and the neural network in a label dependent manner.  This is unlike traditional boosting where the stage weights are fixed and derived based on simple functions of the error of each stage.

LMNs can be used for online adaptation in a variety of settings.  We compare not only with existing MANNs but also state of the art meta-learning methods that retune parameters~\cite{Rei15,Finn2017ModelAgnosticMF}.
On online sequence prediction tasks spanning five applications like user trajectory predictions and next-click prediction,  we show significant gains.
Second, we report higher accuracy in two different settings of the popular Omniglot image classification task. Finally, we present results from two popular language modeling benchmarks and report improvements over state of the art.


\section{Online Model Adaptation using Labeled Memory Networks (LMNs)}
\label{sec-online}
\paragraph{Problem Description}
Online model adaptation kicks in at the time of deploying a batch trained model. Our focus is classification models in this paper.  During deployment the model sees inputs one at a time in sequence.  At a time $t$ for the input $\vx_t$ we predict the label  $\hat{y}_t$,  after which the true output $y_t$ is revealed.  The online model adapter decides how to improve the next input's $\vx_{t+1}$ prediction by combining the limited labeled data $\{(\vx_1,y_1),\ldots,(\vx_t,y_t)\}$ with the batch trained model.  The adaptation has to be fast and performed synchronously at each step.  We are agnostic about how the classifier is batch trained. However, we assume that the adapter can be trained using several such sequences representative of the deployment scenario.
Many problems such as user trajectory prediction, language modeling, and few-shot learning can be cast in such formulation.



We describe our method of online adaptation in two phases.  We first explain how LMNs make predictions for an input $\vx_t$, and next how they adapt when the true label $y_t$ is revealed.  LMNs comprise of three components: the primary classification network (PCN), the memory module, and a combiner network. Our architecture is depicted graphically in \ref{fig:lmn}. We describe each component of the architecture next.

\subsection{Primary Classification Network (PCN)}
This refers to the batch trained neural network that we seek to adapt. 
The PCN may be stateful or stateless.  For example, in applications like trajectory prediction the sequence of inputs are stateful, whereas in tasks like image classification the inputs are stateless. 
Our only assumption is that the last layer of PCN transform each $\vx_t$ into a real vector $\vh_t \in R^d$ before feeding to a softmax layer to predict a distribution over the class labels.  
We use $\vbeta_y \in R^d$ to denote the softmax parameter for class $y$, so the score  $r_{ty}$ for predicting class $y$ for input $x_t$ is
\begin{equation}
\label{eq-pcn}
\begin{split}
 r_{ty} = \frac{\exp(\vbeta_y \vh_t)}{\sum_{z}\exp(\vbeta_z \vh_t)} = \text{softmax}(\vbeta \vh_t)
 \end{split}
\end{equation}
%

\subsection{Memory}
The memory consists of $N$ cells. Each cell $m$  is a 3-tuple with:
$\ell_m$ denoting the label of cell $m$,
$\vM_m$ denoting the hidden vector stored in $m$,
$\util_m$ denoting a weight attached to the cell.  This storage format is similar to what is used in existing MANNs.  But the way in which we read, use, and update the memory is very different.  In existing MANNs memory is viewed as a part of the main network.  Memory values are read based on matching a key with the stored vector $\vM_m$ and the read values are processed by the main network to produce the output.  In contrast we view memory as a learner, specifically an online learner that is only loosely integrated with the PCN. We describe here our alternative design.

We index the memory with the class label $\ell_m$  so that given a label $y$, we can enumerate all cells with label $y$.  The memory provides a score over each class label $y$ for an input $\vx_t$.
We use the PCN last layer output $\vh_t$ as an embedding of $\vx_t$ on the basis of which we can compute the kernel  between $\vh_t$ and a memory vector as  $K(\vh_t,\vM_m) = \exp(\lambda ~\text{cosine}(\vh_t,\vM_m))$.
Given $\vh_t$ and $y$ we read a vector $M_{ty}$ along with a scalar weight $\alpha_{ty}$ calculated as
\begin{equation}
\label{eq-mem-read}
\begin{split}
 M_{ty},\alpha_{ty} = \sum_{m:\ell_m = y} w_{tm}\{\vM_m,\util_m\} \\ 
 w_{tm} =  \frac{K(\vh_t,\vM_m)}{\sum_{m': \ell_{m'} = y} K(\vh_t,\vM_{m'}) }
\end{split}
\end{equation}
This method of reading is very similar to soft addressing used in most memory models.  But the key difference is that we take average only over cells with label $y$, and not all cells.


This label specific vector $M_{ty}$ read from memory is used to get a score for a class $y$ as:
\begin{equation}
\label{eq-mem-score}
\lmm_{ty}  = \frac{\alpha_{ty}^\delta K(\vh_t, M_{ty})} { \sum_{y'} \alpha_{ty'}^\delta K(\vh_t, M_{ty'})}
\end{equation}
where $\delta$ is a strength parameter. The memory can thus be viewed as a kernel classifier with $\alpha_{ty}$ denoting the weight of the kernel.


When memory is large, the time to compute memory scores could be a concern but it is easy to get fast approximate scores using a similarity index~\cite{Guo16,kaiser2017}, and approximate nearest neighbor search has been used for large scale memories \cite{RaeHHDSWGL16,chandarSHPGY16}.

\subsection{Combining memory and PCN}
The final score for a label $y$ is computed by combining pcn-scores $r_{ty}$ and memory scores $\lmm_{ty}$ with an interpolation parameter $\theta$.
We made $\theta$ a dynamic variable, because we expect the relative importance of memory and PCN to evolve as more labeled data becomes available online. Following Adaboost ~\cite{Schapire:1999adaboost}, one way to choose $\theta$ as a function of the two classifier stages. However we obtained better results by using a RNN to determine $\theta$ as a function of label and time as well. The RNN works on a label dependent state and at each step takes three sets of input. First is the input embedding $\vh_t$, second are binary indicators $e^m_{t-1}, e^{pcn}_{t-1}$ which indicate whether the memory and PCN had error at the previous output, and the third set are label probabilities predicted by the memory and PCN. The RNN acts at each step on the state to produce label dependent outputs, from which the relative weights of memory and PCN are obtained via a sigmoid layer.

\begin{equation}
\label{eq-comb}
\begin{split}
 \theta_{ty}& = \sigma(W_\theta \mu_{ty} + b_\theta) \\
 \mu_{ty} &=  \text{RNN}(\mu_{t-1,y}, \{\vh_t, e^{pcn}_{t-1},  e^m_{t-1}, r_{t-1,y}, \lmm_{t-1,y}\}) \\
 \end{split}
\end{equation}

The final predicted distribution over the labels for an input $\vx_t$ is
\begin{equation}
\label{eq:overall}
P_t(y|\vx_t) \propto  (1-\theta_{ty})r_{ty} + \theta_{ty} \lmm_{ty}
\end{equation}

\paragraph{Training the combiner network} The combiner parameters are trained to minimize the loss on $P_t(y|\vx)$ over a labeled sequence of instances that are representative of the deployment setting.
We take a pre-trained PCN and augment it with the memory and combiner modules. This new network is then trained in the online setting, by providing it data in a sequential manner.

An advantage of this architecture is that memory and PCN are loosely coupled, allowing the modules to be plugged in over existing models and can mix at varying levels depending on the amount of per-domain data. We show in the experimental section that even with $\theta_{ty}$ fixed to a single hyper parameter (over all $t,y$), LMNs are competitive with state of the art model adaptation methods.

\begin{figure*}
\begin{subfigure}[b]{0.5\textwidth}
\begin{center}
  \includegraphics[width=0.9\hsize]{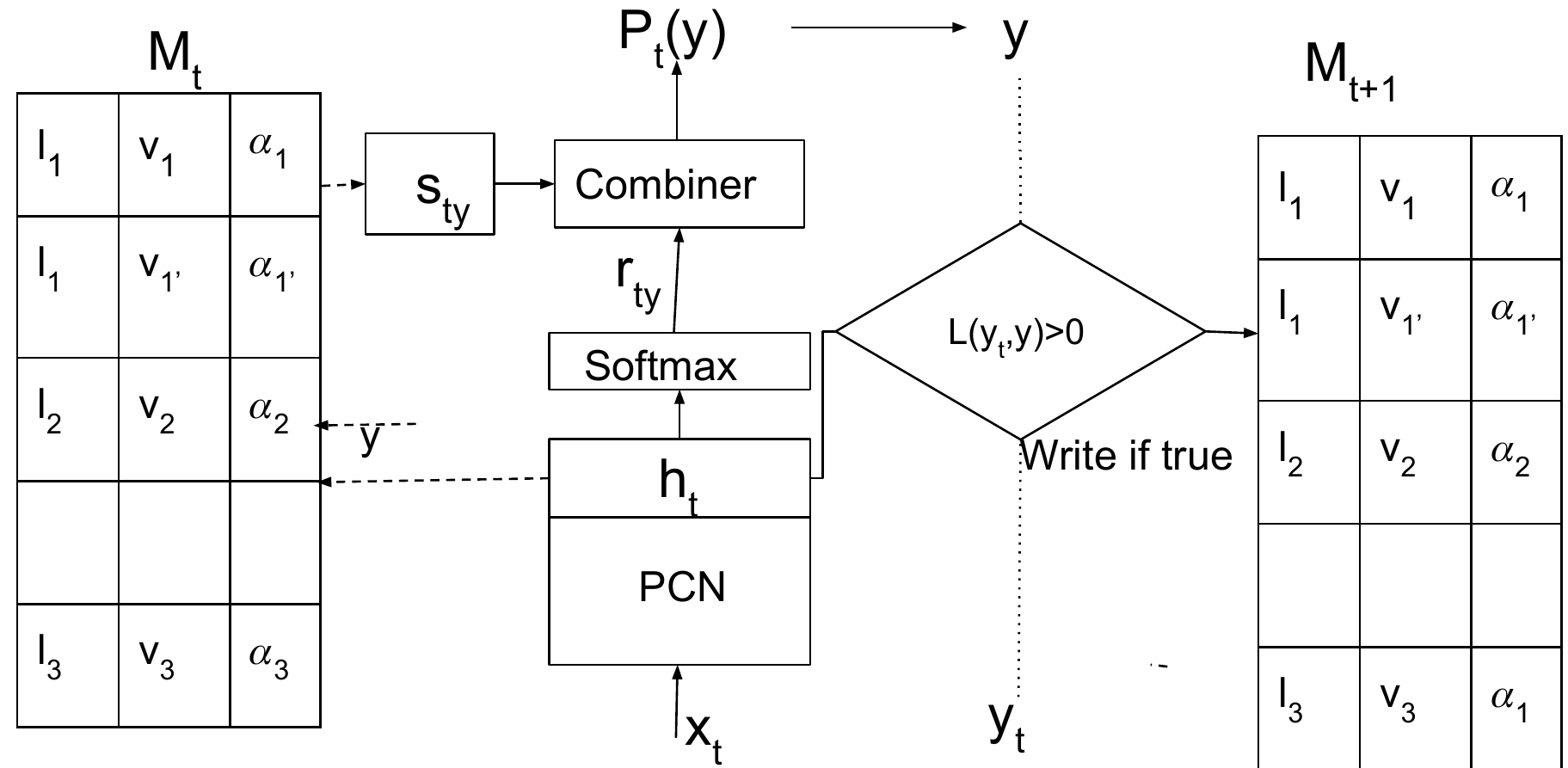}
  \caption{LMN}
  \label{fig:lmn}
  \end{center}
 \end{subfigure}
 \rulesep
 \begin{subfigure}[b]{0.5\textwidth}
 \begin{center}
  \includegraphics[width=0.9\hsize]{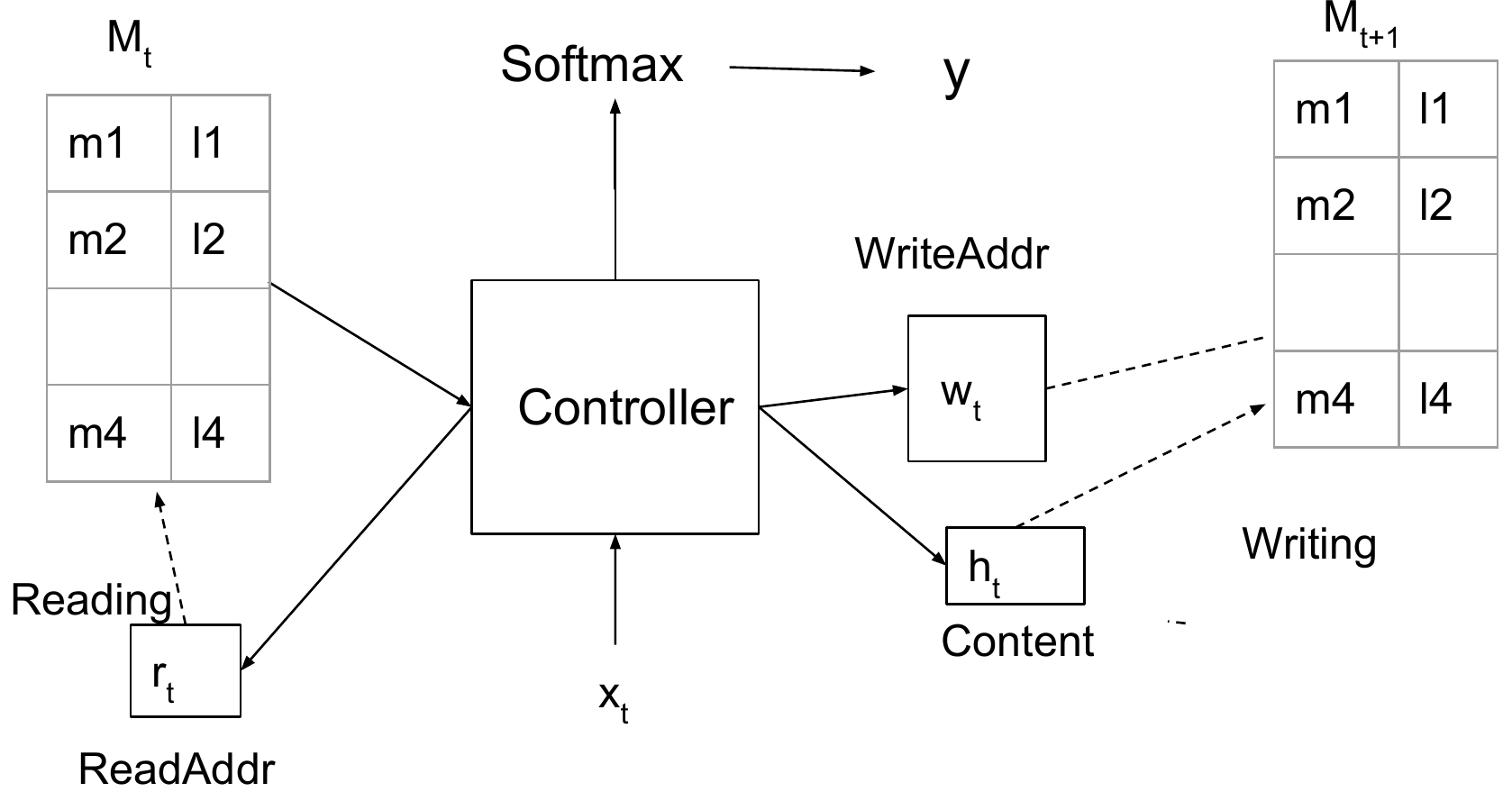}
  \caption{Memory network}
  \label{fig:mem_net}
  \end{center}
 \end{subfigure}
\caption{Comparing Memory Networks and LMN}
\label{fig:net_comparison}
\end{figure*}

\subsection{Adapting to true label $y_t$}
When the true label of $\vx_t$ is revealed we take two steps: update the state of the combiner RNN as discussed in Equation~\ref{eq-comb}, and write to memory if needed.  We describe our method for memory updates next. 

The memory is partitioned across labels, and each memory cell is a tuple of label, content and $\alpha$.  
The way in which we update our memory is consistent with the role that memory serves of being a second classification stage that is updated online.  We next discuss the three parts of our memory write strategy: when to write, how to write, what to replace and why. The overall algorithm is depicted in Figure~\ref{alg-write}.
\paragraph{When to write}
Like in online learners, we update memory only when the margin of separation between the scores of true and closest incorrect label is small. Specifically we update only if $\log P_t(y_t|\vx_t) - \max_{y \ne y_t} \log (y | \vx_t)$ is less than a margin threshold.  For example, instances that are confidently classified by the PCN may never be fed to the memory.     
In contrast,  existing MANNs write to memory for every instance and are prone to fill up the memory on cases that can be accurately handled by PCN.
\paragraph{What to write}
Consider the content $\vM_m$ and weight $\alpha_m$ associated with
each memory cell $m$ at time $t$.   We update the content of cells with label $y_t$ using $\vM_{m} = \vM_{m} + w_{tm}{\vh_t}$ where $w_{tm}$ is the fractional similarity of $\vh_t$ to contents of cell $m$ (Equation~\ref{eq-mem-read}). The weight $\alpha_m$ of the cell is incremented by the fractional similarity $w_{tm}$ to $\vh_t$ after decaying old weights.
Further, if the prediction is incorrect we attempt to create a new cell in memory  by replacing an existing cell.  
\paragraph{What cell to replace} We replace the cell with the smallest weight among all cells with the same label as $y_t$. 
Since we update $\alpha_m$ based on its fractional similarity to instances that had non-zero loss, this method essentially replaces the cell that is least useful for defining the classification task.  Unlike existing MANNs that replace a cell least recently used across all labels, we replace only among cells with label $y_t$. A pure LRU based replacement could easily lead to forgetting of rare classes.  Our method of replacement enables us to implement a fairer classifier atop the memory. 

\begin{figure}
\begin{center}
\begin{algorithmic}
  \STATE {\bf Input:} $y_t, P_t(y|\vx_t), \vh_t, M_t = \{(\ell_m, \vh_m, \alpha_m)\}$
  \STATE $\hat{y} = \argmax_y P_t(y|\vx_t)$
  \STATE $\tilde{y} = \argmax_{y \neq y_t} P_t(y|\vx_t)$
  \STATE $\loss_t = - \min(0, \log P_t(y_t|\vx_t) - \log P_t(\tilde{y}|\vx_t) - \text{margin} )$
  \STATE If $\loss_t=0$, no update. Return.
  \STATE New cell $C$: $(\ell_{y_t}, \vh_t, 1)$
  \STATE $j = \argmin_{m:\ell_m = y_t} \alpha_m$
  \STATE $\vw_t = \text{softmax}\{\text{Cosine}(\vh_t^T \vM_m) :\ell_m = y_t\}$
  \STATE $\vM_{m} = \vM_{m} + w_{tm}{\vh_t}~~~~\forall m: \ell_m = y_t$
  \STATE $\alpha_{m} = \alpha_m*\text{decay} + w_{tm}~~~~\forall m: \ell_m = y_t$
  \IF {$y_t \ne \hat{y}$ and $|m:\ell_m=y_t| > 1$}

  \STATE Replace cell $j$ with new cell $C$.
  \ENDIF
\end{algorithmic}
\caption{\label{alg-write}Memory updates in Labeled memory network. In our experiments we used a decay value of 0.99.   The margin is a hyper-parameter.}
\end{center}
\end{figure}

Our memory updates are analogous to gradient updates in an online SVM learner. However we operate in a limited memory setting and merge and delete support vectors similar to budgeted-PEGASOS \cite{WangCV10}. If memory vectors are considered as SVM parameters, then the gradient w.r.t these parameters of the objective is precisely the current vector $\bf{h}$. In an online SVM learner when the loss is non-zero, the input element is added to the set of support vectors. Similarly in LMN, when margin loss is non-zero, the current vector $\bf{h}$ is added to the memory. The contribution of this specific memory update for the memory scores in subsequent steps will then be proportional to $\alpha_{ty}^\delta K(\vh_t, \bf{h})$. This acts as an online-SVM learner in the dual form, where the $\bf{h}$ are the support vectors, and $\alpha$ act as the associated dual variables. 

\section{Related Work}

\paragraph{Memory in neural networks}
The earliest example of the use of memory in neural networks is attention~\shortcite{BahdanauCB14}.
Attention as memory is slow for long histories, leading to the development of several more flexible memory-based architectures~\cite{Weston16}.
Neural Turing Machines (NTMs)~\cite{GravesNTM} were developed for end-to-end learning of algorithmic tasks. One such task where NTMs were shown to work was learning N-gram distribution from token sequences. Since this is related to online sequence prediction, our first model was based on NTMs.  However, on our real datasets we found NTMs to not be very effective. The reasons perhaps is the controller's difficulty with adaptively generating keys and values for memory operations.  Dynamic-NTMs (DNTMs)~\cite{GulcehreCCB16} alleviate this via fixed trainable keys, and are shown to aid QA tasks but they did not work either as we will show in our experimental section.
Like LMNs, DNTMs also propose discrete memory addressing but the keys are trained from input $\vx_t$ unlike in LMNs where the discrete key is class label that requires no training.
Another difference with LMNs is that the memory is very tightly integrated with the neural network and requires joint training. 


\paragraph{MANNs for classification}
More recent MANNs designed for classification employ a looser coupling~\cite{kaiser2017,SantoroBBWL16} and store hidden vectors and discrete labels as memory contents as in LMNs. 
However, they differ in how the memory is addressed, updated, and used (shown graphically in Figure \ref{fig:net_comparison}).
First, LMNs treat memory as an on-line learner and update memory only when loss is non zero unlike all previous MANNs that update the memory for every step.  This allows memory to be used more effectively to store instances when PCN is weak.
Second, in MANNs memory reads are fed back into the main model for getting a prediction.  Instead, in LMNs memory reads are loosely combined with the PCN scores allowing us to use memory for plug-and-play adaptation of a trained model.  Finally, most MANNs use a global LRU replacement strategy whereas we replace only within in a label based on importance. This reduces the bias against rare classes present in LRU policies.



%
\paragraph{Model Adaptation by parameter re-learning}
Another way to adapt is by training or tuning parameters. The methods of \cite{Rei15} and \cite{Huang2015MaximumAP} train only a subset of parameters that are local to each sequence.  More recently,\cite{Finn2017ModelAgnosticMF} and \cite{ravi17} propose meta-learners that "learn to learn" via 
the loss gradient. In general, however such model retraining techniques are resource-intensive. In our empirical evaluation we found these methods to be slower and less accurate than LMNs. 

\paragraph{Online Learning} Online learning techniques such as \cite{shalevshwartz:icml07} for learning kernel coefficients is relevant if we view the memory vectors $ \vh_m$ acting as the support vectors and the memory scalars $\alpha_m$ as the associated dual variables.  Our setup is a little different in that we employ a mix of batch and online learning. Our proposed scheme of memory updates and merge was inspired by the gradient updates in PEGASOS, and in the case of exactly one cell per label reduces to a specific form of budgeted-PEGASOS\cite{WangCV10}.



\section{Experiments}
We next present empirical comparison of LMNs with recent MANN based meta-learners and state of the art parameter re-learning based adaptive models. We experiment \footnote{code to be available on https://github.com/sshivs/LMN} on three different supervised learning tasks that require online model adaptation: sequence prediction on five datasets spanning applications like user trajectory prediction and next click prediction,  image classification under online addition of new image labels, and language modeling.

\subsection{Online sequence prediction}
\label{sec-expt-online}
In this task the data consists of a sequence of discrete tokens $x_1,\ldots,x_n$ and the label $y_t$ to be predicted at time $t$  is just the next token $x_{t+1}$.  The inputs are stateful and therefore the PCN has to be a RNN.
We collected five such datasets from different real-life applications. In Table~\ref{sequence-data} we summarize the average length of each sequence, number of tokens, and the number of sequences in the training and test set.  
\begin{itemize}
\item
FSQNYC and FSQTokyo are Location Based Social Network data collected by ~\cite{yang2014} from FourSquare of user check-in at various venues over an year.

\item Brightkite \cite{ChoMyLes11} is a user check-in dataset made available as part of Stanford Network Analysis Project~\cite{snapnets}.

\item
Geolife \cite{geolife09} is the trajectory data of people collected over multiple days, and provides the GPS coordinates of people tracked at almost a minute interval. We discretize the locations with a resolution of 100 meters and limit to the densest subset around the city.
\item
The Yoochoose dataset \cite{Ben-Shimon:2015:RCY:2792838.2798723} is the click event sessions for a major European e-tailer collected over six months. 

\end{itemize}

\begin{table}
\begin{center}
\begin{tabular}{|l|r|r|r|r|} \hline
Name & No. of train & No of test & Avg seq. & \# Tokens \\
         &  sequences & sequences & length &  or labels  \\ \hline
Brightkite & 1800  & 521  & 238  &  22811 \\
FSQNYC & 670  &  264  & 90  &  8325 \\
FSQTokyo & 1555  & 672  & 160  &  12589 \\
Geolife & 220  & 20  & 8000 &   31273\\
Yoochoose &  450523 & 112279  & 13 &   39481\\ \hline
\end{tabular}
\end{center}
\caption{\label{sequence-data} Statistics of data used for on-line sequence learning}
\end{table}
\paragraph*{Experimental setups}
In all experiments we used the Adam optimizer~\cite{KingmaB14}.  The PCN is a GRU and the input is the embedding of the true observed token $y_{t-1}$ at the previous time.  
The number of memory cells is equal to the number of labels.


\paragraph*{Methods compared}

We evaluate these tasks on six different methods.
\begin{itemize}
\item
As a baseline we train a larger LSTM \cite{Hochreiter97lstm} which has roughly 5 times more RNN parameters compared to the PCN used in LMN.  This lets us evaluate if a larger LSTM state could substitute for memory.
\item
Next we choose two recent approaches from the family of meta-learners that re-tune model parameters: the method of \cite{Rei15} since it was specifically proposed for sequence prediction and the more recent but generic method MAML of \cite{Finn2017ModelAgnosticMF}. Both these models used an LSTM of size 50 as the base learner. After each true label is encountered these models compute gradient of the loss with respect to the adapting parameters, and apply those updates, before processing the next input.

\item
As a representative of MANNs, we implemented a version of D-NTM \cite{GulcehreCCB16} where we made two changes to adapt to the on-line prediction task. First, we use the previous read address as the new write address, and second we derive the new content from read memory content and $y_{t-1}$. We tried several other tweaks, including the unchanged D-NTM and obtained best results with these changes.

\item We compare these with LMN. To illustrate the importance of adaptively reweighting the PCN and memory scores as per Equation~\ref{eq-comb}, we also tried another model called LMN-fixed.  In LMN-fixed $\theta_{ty}$ is fixed for all $t$ and $y$ and is determined by batched hyper-parameter optimization.
\end{itemize}

\begin{table*}[!htb]
\begin{center}
\begin{tabular}{|l|r|r|r|r|r|r|r|} \hline
  & Baseline & \multicolumn{2}{|c|}{Parameter-retune} & \multicolumn{3}{|c|}{Memory-based} \\ \hline
Name &  LSTM & Rei &  MAML & DNTM & LMN-fixed & LMN \\ \hline
Brightkite  & 10.7 (0.11) & 10.01 (0.18) &   4.27 (0.47) & 9.63 (0.13)& 3.88 (0.51) & \textbf{3.57 (0.55)}\\
FSQNYC  &  8.95 (0.03) & 8.72 (0.07)   & 6.55 (0.16) & 9.01 (0.05)& 6.13 (0.25) &\textbf{5.54 (0.27)}\\ 
FSQTokyo & 8.14 (0.08)  &  6.95 (0.13)&   5.68 (0.23) & 7.25 (0.11) & 5.40 (0.26) & \textbf{5.32 (0.28)} \\ 
Geolife  &1.13 \textbf{(0.84)}  &  1.08 (0.83)&  - & 1.11 (0.82) & \textbf{1.05 } (0.83) &1.08 (0.83) \\ 
Yoochoose & 5.01 (0.24) & 5.05 (0.23)&   - & 5.01 (0.23) & 5.01 (0.24)  &\textbf{4.96 (0.25)}\\ \hline
\end{tabular}
\end{center}
\caption{\label{sequence-results} Log Perplexity and MRR(in parantheses) on online sequence prediction tasks}

\end{table*}

\begin{table*}[!htb]
\begin{center}
\begin{tabular}{|l|r|r|r|r|r|r|r|} \hline

\end{tabular}
\end{center}
\end{table*}
{ \bf Results}
In Table \ref{sequence-results} we report the log-perplexity and mean reciprocal rank (MRR) of all six methods on the five datasets. We make the following observations from these.
\begin{enumerate}
\item LMN-based online model adaptation significantly boosts accuracy over a baseline LSTM with five times larger state-space.  
The datasets vary significantly in their baseline and LMN accuracies. FSQNYC, FSQTokyo and Brightkite datasets have MRR increasing by 100 and 400\%.  For Yoochoose improvements are smallest because most sequences are very short (13 on average). In Geolife the sequences are much larger but the baseline MRR (0.84) is already high indicating a saturation point.
\item The improved accuracy of LMN over LMN-fixed illustrates the importance of online reweighting of PCN and memory. We observe improvements over baseline on almost all datasets, even on Yoochoose where none of the other methods did.  However, LMN-fixed is better than parameter retuning and DNTMs establishing that even without training the combiner, LMNs can be used as plug-and-play model adapters.
\item Meta learners that retune parameters for adaptation are indeed effective compared to the baseline.  The Rei method that retunes only a designated 'sequence-vector' parameter is less effective than the MAML that retunes all parameters. Yet, our proposed LMN, (or its simpler LMN-fixed variant) provides an even greater boost.  For example, for Brightkite the MRR increases from 0.11 to 0.18 with Rei, to 0.47 with MAML, and to 0.51 with LMN-fixed and 0.55 with LMN. 
A major shortcoming of MAML is that training the meta-learner is very slow.  We were not able to complete the training of MAML on our two largest datasets Geolife and Yoochoose within a reasonable time. In contrast, LMNs are significantly faster as they do minimal re-training.
DNTM, another MANN-based approach is not as effective as even LMN-fixed and we believe it is mainly because of the differences in their respective memory organization. 
\end{enumerate}
These experiments establish that a well-designed MANN is a practical option for accurate and efficient online adaptation for next-token prediction tasks.  

\subsection{Online Adaptation of Image Classifiers}
\label{sec-expt-rare}
We next demonstrate online adaptation of an image classifier to new labels observed during testing.  Unlike existing work in few-shot learning~\cite{VinyalsBLKW16} where each test sequence has its own independent prediction space, we consider the more useful and challenging task where new labels co-exist with labels seen in the batch-trained PCN.  

\paragraph*{Dataset and setup} We use the popular omniglot dataset~\cite{Lake1332}. The base data-set consisted of 1623 hand-drawn characters selected from 50 different alphabets.  \cite{VinyalsBLKW16} extended the base data-set by various rotations of the images, and this increases the number of categories to 4515.
 We create an online variant of this dataset as follows. We arbitrarily select 100 classes as unseen and 250 as seen classes. During training only seen classes are provided, but the test data has a uniform mix of all classes. Each test sequence is obtained by first randomly selecting 5 labels from the 350, and then choosing different input images from the selected labels up to a length of 10.
Thus, in each sequence we expect to encounter $\frac{10}{7}$ new labels on average.   Accuracy is measured only over second occurrence of each of the five labels much like in one-shot learning experiments. The one major difference is that in our case the prediction space is all the seen labels and the expected new labels, whereas in few-shot learning the prediction space is only the 5 labels chosen for that test sequence. The results are averaged over 100 sequences. 

\paragraph{Methods compared} We compare results of LMN with MAML~\cite{Finn2017ModelAgnosticMF} the most recent meta-learner that can work in this setup. We do not compare with the two recent MANNs~\shortcite{kaiser2017,SantoroBBWL16}  that report Omniglot results on few-shot learning because their techniques do not easily extend to the case where new labels share label space with a pre-trained classifier. For reference we also report accuracy on the baseline classifier that will certainly mis-classify examples from the new class. We use as PCN a convolutional network, with the same configuration as used in ~\shortcite{kaiser2017}.  

.  
\begin{table}[!htb]
      \centering
        \begin{tabular}{|l|r|r|r|} \hline
Model & Overall & New labels & Seen labels \\ \hline
Baseline & 45\% & 0 & 63\% \\
MAML & 56\% & 49\% & 59\%\\ 
LMN  & {\bf 86\%} & {\bf 71\%} & {\bf 92\%} \\  \hline 
\end{tabular}
 \caption{\label{tab-unk}Accuracy on Online adaptation for Omniglot}
\end{table}
\paragraph{Results} 
In Table~\ref{tab-unk} we report our results.  The baseline that does no adaptation achieves an overall accuracy of 45\% while obtaining an expected accuracy of 0\% on the new labels and $63\%$ on the seen labels.  MAML boosts the overall accuracy to 56\%, while obtaining an accuracy of 49\% on the new labels.  However, compared to the baseline the accuracy on seen labels has dropped. This is similar to catastrophic forgetting \shortcite{French99} and may be because during meta-learning updates the weights of the shared representation layer deteriorate.
The LMN architecture is better able to absorb new labels and changing priors of existing labels.  LMN achieves an overall accuracy of 86\% with 71\% on new labels. The accuracy increases for the seen labels as well because LMN is able to use the memory for storing prototypes of examples with non-zero loss even from seen labels. Thereafter, the combiner RNN can pay more heed to the labels seen during the adaptation phase.


\subsubsection{Plain few-shot learning}
To enable comparison with the few-shot results of many recent published work, we also report results of LMN on this task.
In few-shot learning, the PCN only generates the input embedding $\vh_t$ and does not assign label scores since each test sequence has its own label space. The memory uses the embedding to assign label scores via Equation~\ref{eq-mem-score} based on which prediction is made without involving any combiner either.
These experiments will compare LMN's memory organization and update strategies with state-of-the-art MANNs that have reported results on few shot learning~\cite{kaiser2017}.  

Our setup is the same as in recent published work ~\shortcite{VinyalsBLKW16}  on few-shot learning but where the total label space has 4K possibilities labels.  This is more interesting and realistic than 5-way classification where many recent methods, including ours, report more than 98\% accuracy with just 1-shot.

We present results with changing memory size. We run the experiments with a memory size equal to T-cell per label (T=2,3..).  We observe that LMN accuracy is much higher than existing MANNs particularly when memory is limited. For example, at 2-cells per label (roughly 8K memory size) we obtain more than 4\% higher accuracy in 1-shot, 2-shot, and 3-shot learning. This proves the superior use of memory achieved via our labeled memory organization

For reference we also compare with parameter retuning-based approach (MAML)~\shortcite{Finn2017ModelAgnosticMF}.  Even though this work reports state-of-art accuracy on 5-way few-shot learning, for 4k way few-shot learning it is not as effective.  A softmax layer that has to arbitrate among 4k new classes perhaps need lot more gradient updates than learnable by meta-learners.


\begin{table}[!htb]
\begin{center}
\begin{tabular}{|l|r|r|r|} \hline
Name &   1 shot &   2 shot &  3 shot \\ \hline
Kaiser (2-cell/label) & 48.2\%	&	58.0\%	&	60.3\% \\ 
Our model (2-cell/label) & \textbf{52.6\%}	&	\textbf{62.5\%}	&	\textbf{64.1\%} \\ \hline
Kaiser (3-cell/label) & 49.7\%	&	60.1\%	&	63.8\% \\ 
Our model (3-cell/label) & \textbf{52.8\%}	&	\textbf{63.0\%}	&	\textbf{67.0\%} \\ \hline
Kaiser (5-cell/label) & 52.7\%	&	63.0\%	&	66.3\% \\ 
Our model (5-cell/label) & \textbf{54.3\%}	&	\textbf{63.2\%}	&	\textbf{66.9\%} \\ \hline
MAML & 44.2\% & 46.5\% & 47.3\%  \\ \hline
\end{tabular}
\end{center}
\caption{\label{fewshot-results1} Test accuracies for 4k way few shot learning}

\end{table}

\subsection{Language Modeling}
Language modeling is the task of learning the probability distribution of words over texts. When framed as modeling the probability distribution of words conditioning on previous text, this is just another online sequence prediction task. The natural dependence on history in this task provides for another use-case of memory. Memory based models have been shown to get improvements over standard RNN based language models ~\cite{SukhbaatarSWF15,merity2016pointer,grave2016improving}.
In the same spirit, we apply LMNs to this task by taking the PCN as a RNN.

We compare our model directly with the recently published neural cache model of ~\cite{grave2016improving} and pointer LSTM of ~\cite{merity2016pointer}. These showcase variant uses of memory to improve the prediction of words that repeat in long text.  The baseline model (and PCN) is an LSTM with same parameters as in \cite{grave2016improving}.
%
We compared on common language datasets Wikitext2 and Text8 with memory sizes 100 and 2000 as used in the previously published work. We used standard SGD as the optimizer. 

 In Table~\ref{tab-lm} we report the perplexities we obtained with LMN along with the results reported in published work for other three approaches. As the table demonstrates we achieve state of the art results in Text8 and Wikitext2. In LMN the auto-modulation caused by considering only cases where the PCN is weak is superior to memory cache. This shows up in tests when memory is constrained, when the focus on mis-predicted outputs in LMN allows for boosted recall, efficient memory utilization, and capturing longer contexts, compared to other models.

\begin{table}
\begin{center}
\begin{tabular}{|l|r|r|} \hline
Name &    Wikitext2 (100 ) &  Text8 (2000)\\ \hline
LSTM & 99.7 & 120.9 \\
Pointer LSTM &	80.8	& -\\ 
Neural cache &  81.6 &  99.9\\
LMN &	\textbf{77.6}	& \textbf{91.1}\\ \hline
\end{tabular}
\end{center}
\caption{ \label{tab-lm} Test perplexity for language modeling on Wikitext and Text8 with memory 100 and 2000 respectively.}  
\end{table}

\section{Conclusion}
We extended standard memory models with a label addressable memory module and an adaptive weighting mechanism for on-line model adaptation. 
LMNs mix limited data with pre-trained models by combining ideas from boosting and online kernel learning while tapping deep networks to learn representations and RNNs to model the evolving roles of memory and pre-trained models.  
%
LMNs have some similarities to recent MANNs but has significant differences.
First, we have a label addressable memory instead of content based addressing. Second, we use memory to only store content on which primary network is weak. 
Third, our model has a loose coupling between memory and network, and hence our model can be used to augment pre-trained models at a very low cost. 
Fourth we use an adaptive reweighing mechanism to modulate the contribution of memory and PCN. 
This LMN is demonstrated to be extremely successful on a variety of challenging classification tasks which required fast adaptation to input and handling non-local dependencies. 
An interesting extension of LMNs is organizing the memory not just based on discrete labels but on learned multi-variate embeddings of labels 
thereby paving the way for greater sharing among labels.

\paragraph{Acknowledgements}
We gratefully acknowledge the support of NVIDIA Corporation with the donation of the Titan X Pascal GPU used for this research.

\bibliographystyle{aaai}
\bibliography{neural-memory.bib}

\end{document}